\documentclass[conference]{IEEEtran}
\IEEEoverridecommandlockouts
\usepackage{cite}
\usepackage{amsmath,amssymb,amsfonts}
\usepackage{algorithmic}
\usepackage{graphicx}
\usepackage{textcomp}
\usepackage{xcolor}
\usepackage{multirow}
\def\BibTeX{{\rm B\kern-.05em{\sc i\kern-.025em b}\kern-.08em
    T\kern-.1667em\lower.7ex\hbox{E}\kern-.125emX}}
\begin{document}

\title{Enabling Smartphone-based Estimation of Heart Rate\\
}


\author{\IEEEauthorblockN{Nutta Homdee\textsuperscript{$1$,$2$}, 
Mehdi Boukhechba\textsuperscript{$1$,$3$}, 
Yixue W. Feng\textsuperscript{$4$}, 
Natalie Kramer\textsuperscript{$5$}, \\
John Lach\textsuperscript{$1$,$2$},
Laura E. Barnes\textsuperscript{$1$,$3$}}
\IEEEauthorblockA{\textit{Link Lab\textsuperscript{$1$}}}
\IEEEauthorblockA{\textit{Department of Electrical and Computer Engineering\textsuperscript{$2$}, 
Department of Engineering Systems and Environment\textsuperscript{$3$}}} 
\IEEEauthorblockA{\textit{Department of Computer Science\textsuperscript{$4$},
Sports Medicine\textsuperscript{$5$}
}}
\textit{University of Virginia}\\
Charlottesville, VA USA \\
{nh4ar, mob3f, yf8vt, 
nak5dy, jlach, lb3dp}@virginia.edu}

\maketitle

\begin{abstract}
Continuous, ubiquitous monitoring through wearable sensors has the potential to collect useful information about users' context.  Heart rate is an important physiologic measure used in a wide variety of applications, such as fitness tracking and health monitoring. However, wearable sensors that monitor heart rate, such as smartwatches and electrocardiogram (ECG) patches, can have gaps in their data streams because of technical issues (e.g., bad wireless channels, battery depletion, etc.) or user-related reasons (e.g. motion artifacts, user compliance, etc.). The ability to use other available sensor data (e.g., smartphone data) to estimate missing heart rate readings is useful to cope with any such gaps, thus improving data quality and continuity. In this paper, we test the feasibility of estimating raw heart rate using smartphone sensor data. Using data generated by 12 participants in a one-week study period, we were able to build both personalized and generalized models using regression, SVM, and random forest algorithms. All three algorithms outperformed the baseline moving-average interpolation method for both personalized and generalized settings. Moreover, our findings suggest that personalized models outperformed the generalized models, which speaks to the importance of considering personal physiology, behavior, and life style in the estimation of heart rate. The promising results provide preliminary evidence of the feasibility of combining smartphone sensor data with wearable sensor data for continuous heart rate monitoring. 
\end{abstract}


\section{Introduction}
Sensor data from mobile phones and wearable devices, such as smartwatches, have the ability to capture continuous \textit{in-situ} data about users' behaviors and wellbeing. One of the unique streams that is passively collected by wrist-worn devices is heart rate. Heart rate is a physiologic measurement that can be useful in monitoring both physical and psychological activity \cite{b1}, \cite{b2}. This data can be further exploited to extract meaningful biomarkers used to understand human health, such as anxiety and stress \cite{b3}, \cite{b4}. It can also provide insight into aspects of disease processes like loss of mobility or decreased physical activity.  

One of the major issues of wearable devices used \textit{in the wild} is missing data.  The gaps in the data stream can be the result of bad wireless channels, motion artifacts, battery depletion, software/firmware errors and updates, and the user simply not wearing the device \cite{b5}. Similar gaps can occur in mobile phone sensor data, but they may not temporally overlap with gaps in the wearable data. The central theme in this work is therefore to explore the relationship between sensor data from the mobile phones and wearable devices to determine if one can compensate for the other during data gaps to preserve continuous monitoring. More specifically, the main research question is to investigate if mobile phone sensor data can estimate heart rate as collected by a wrist-worn wearable device, such as a smartwatch.



Previous works have researched the ability of phone sensors to estimate heart rate. In work by Herandez, McDuff, and Picard \cite{b6}, they estimate heart rate using a wrist-worn accelerometer and gyroscope during sleep. In this work, the authors use a controlled experiment, varying phone positions and location and reported a mean absolute error of 1.16 beats per minute (STD: 3). The phone positions included standing up, sitting down, and lying down, which were repeated before and after exercising, followed by watching a video, listening on the phone, and browsing the Internet while sitting down \cite{b7}. Another work \cite{predtHR} proposed a smartphone-based system for monitoring walking workload (estimating heart rate). This work uses machine learning techniques to estimate heart rate from the acceleration and speed (GPS data) during walking.

While these projects show a relationship between smartphone sensor data and heart rate collected by a smartwatch, these studies were conducted in a controlled setting, such as asking the participants to carry the phone in specific positions and perform specific activities (e.g., walking). Furthermore, these works only focus on analyzing smartphone motion and smartwatch heart rate, excluding other smartphone available sensors.


In this paper, we describe uncontrolled data collections and data analytics to correlate smartwatch heart rate data with a variety of smartphone sensor data, including motion, GPS, time of day, and other contextual information. With such correlations, it becomes possible to estimate heart rate continuously from just smartphone data. While such estimates would not be as accurate as heart rate data from a smartwatch, it could be used to fill gaps when smartwatch data is not available.

\section{Study Design}
Twelve university students volunteered to participate in a seven-day study period to collect both wrist-worn sensor data and phone sensor data. The participants provided a relatively homogeneous sample mitigating the impact of confounding factors. Both Android and iOS phones were being used among the group. A customized mobile app (Sensus) [8] was installed on participants’ personal smartphones and was programmed to collect GPS coordinates every 15 second and accelerometer data at 1 Hz.  All data were transmitted wirelessly to a secure Amazon Web Server for further analysis (see Figure 1). 

\begin{figure}[htbp]
\centering
\includegraphics[width=8.5cm]{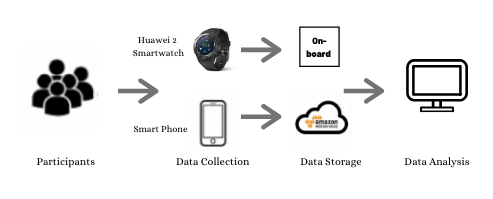}
\caption{Data collection process}
\label{studyDesign}
\end{figure}

Participants were also given a smartwatch (Huawei Watch 2) to wear. The watch recorded optical heart rate (beats per minute) at a sampling rate of 1 Hz. Participants were asked to wear the watch as much as possible during the 7 days. Data was stored on-board and extracted at the end of the collection period for further analysis.

\section{Methods}

\subsection{Sensor Data}
\begin{itemize}
\item Accelerometer Data: Accelerometry data is used to asses physical activity level of the participant. The Sensus application collects 3-axis, x,y, and z accelerometer data. we first calculated the magnitude of acceleration:
\begin{equation}
Acceleration Magnitude=\sqrt{x^2+y^2+z^2}
\end{equation}
Since the phone was used in the participants’ natural environment, the orientation of the phone could affect the accelerometer magnitude. To have an orientation free motion feature, the magnitude value of 1.0 meant the phone was still; to make this value more meaningful for prediction models, we decided to create features for each of the accelerometer metrics that represented the deviation from 1.0. In the interest of maintaining as much data as possible, all accelerometer metrics (x, y, z, and magnitude) were used for analysis.
\item GPS Data: GPS data provide information about participants’ mobility patterns. We create six location data features, three latitude features, and three longitude features, by rounding the latitude and longitude digits. Less decimal digits of the GPS data means it covers more area, and the more decimal digits gives a more detailed location such as different buildings in the  university campus. The GPS features are latitude and longitude data with 2-decimal digits, 1-decimal digit and no decimal digit.
\item Time Feature: Time of the day can be indicative of fluctuations in heart rate. For instance, heart rate during sleep is known to be lower than when performing other activities \cite{sleep}. Datetime was parsed into hours, minutes and seconds and added to the feature space as well.
\item Heart Rate: The heart rate data was collected using the smartwatch. For personalized prediction models, each model was trained and tested on the same subject using raw heart rate readings. To compare heart rates across participants, we used general models. However, each participant's heart rate can depends on the person's baseline physiological such as age or fitness. Thus, using the beat-per-minute heart rate may skewed the result as the heart rate of a much older person may response differently to the same physical activity to the younger person. It made the most sense to standardize heart rates across the participants to remove baseline physiological differences. Thus, z-scored heart rate was utilized in the generalized model.
\end{itemize}

\subsection{Predictive Models}

1)	Personalized Models: We used three algorithms to test the predictability of heart rate: ridge regression, support vector machine (SVM), and random forest decision tree. We chose to start with regression as a natural first step in analyzing the data and ultimately chose to use the ridge regression analysis in order to diminish the multicollinearity of the variables. In addition to the regression model, the SVM regression model was chosen due to its ability to handle non-linear data. Lastly, to better understand the results of the SVM model, a random forest was utilized. 
2)  General Models: In addition to our personalized models, we also created general models using all four participants' data. To accommodate for the different ranges of heart rate baseline of each participant, the models are targeted the heart rate's z-score instead of the actual beat-per-minute heart rate. The same parameters (e.g. alpha, epsilon, C, number of the tree) as the personalized models were used.
The included features for the predictive models are listed in Table 1.
\begin{table}[htbp]
\centering
\setlength\tabcolsep{3pt}
\caption{Data Features and Prediction Models Summary}
\begin{tabular}{|l|l|l|l|l|l|l|}
\hline
\multirow{2}{*}{Feature} & \multicolumn{3}{l|}{\begin{tabular}[c]{@{}l@{}}Individual Models\\ (target: heart rate (BPM))\end{tabular}}                  & \multicolumn{3}{l|}{\begin{tabular}[c]{@{}l@{}}General Models\\ (target: z-score heart rate)\end{tabular}}                   \\ \cline{2-7} 
                         & \begin{tabular}[c]{@{}l@{}}Ridge \\ Regression\end{tabular} & SVM & \begin{tabular}[c]{@{}l@{}}Random \\ Forest\end{tabular} & \begin{tabular}[c]{@{}l@{}}Ridge \\ Regression\end{tabular} & SVM & \begin{tabular}[c]{@{}l@{}}Random \\ Forest\end{tabular} \\ \hline
\multicolumn{7}{|c|}{Accelerometer Features}                                                                                                                                                                                                                                           \\ \hline
X                        & \multicolumn{6}{l|}{X-axis acceleration}                                                                                                                                                                                                                    \\ \hline
Y                        & \multicolumn{6}{l|}{Y-axis acceleration}                                                                                                                                                                                                                    \\ \hline
Z                        & \multicolumn{6}{l|}{Z-axis acceleration}                                                                                                                                                                                                                    \\ \hline
Magnitude                & \multicolumn{6}{l|}{Total acceleration magnitude (equation. 1)}                                                                                                                                                                                             \\ \hline
\multicolumn{7}{|c|}{Location Feature (Latitude and Longitude)}                                                                                                                                                                                                                        \\ \hline
2-decimal                & \multicolumn{6}{l|}{2 decimal digit latitude and longitude}                                                                                                                                                                                                 \\ \hline
1-decimal                & \multicolumn{6}{l|}{1 decimal digit latitude and longitude}                                                                                                                                                                                                 \\ \hline
0-decimal                & \multicolumn{6}{l|}{0 decimal digit latitude and longitude}                                                                                                                                                                                                 \\ \hline
\multicolumn{7}{|c|}{Time Feature}                                                                                                                                                                                                                                                     \\ \hline
Hour                     & \multicolumn{6}{l|}{hour of the day (0-23)}                                                                                                                                                                                                                 \\ \hline
Minute                   & \multicolumn{6}{l|}{minute of the day (0-59)}                                                                                                                                                                                                               \\ \hline
Second                   & \multicolumn{6}{l|}{second of the day (0-59)}                                                                                                                                                                                                               \\ \hline
\end{tabular}
\end{table}

\section{Results}
In this section, the performance of each regression models is validated with 5-fold cross-validation setting. Then we calculate 1.) The coefficient of determination or the R squared value and 2.) Root mean-squared error (RMSE). Both the R-squared value and RMSE are averaged among all participants. Our approach was compared with a baseline model that uses a moving-average interpolation method. The baseline uses the thirty minutes prior heart rate to interpolate the next thirty minutes.

\subsection{Models Results}
1)	Personalized Models: The personalized models result is shown in Figure 2. The r-squared and RMSE are shown for all models, baseline (30-min interpolation), Ridge regression, SVM, and random forest. We obtain the r-squared value of 0.28, 0.41, 0.36, and 0.82 and the RMSE value of 21.88, 8.79, 11.02, and 5.06 for the baseline, Ridge regression, SVM, and random forest models respectively. Results suggest that all three predictive models outperformed the baseline interpolation model. Which means that the three predictive models were able to learn the association between the smartphone sensor data and heart rate. Furthermore, Random Forests showed the best performance in terms of r-squared and RMSE. Example of heart rate prediction using Random Forest as compared with ground truth is shown in Figure 3.

\begin{figure}[htbp]
\centering
\includegraphics[width=8.5cm]{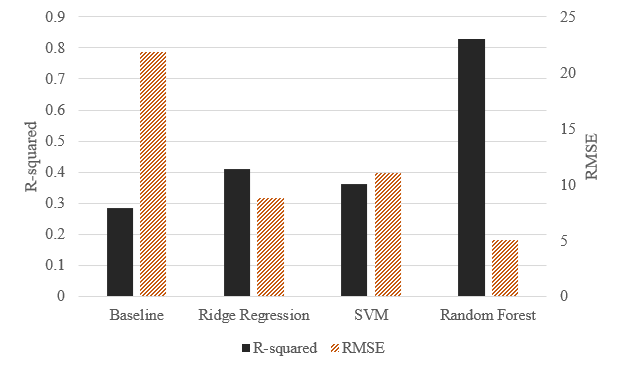}
\caption{R-squared and RMSE results of personalized ridge regression, SVM, and random forests as compared with a baseline moving-average interpolation method.}
\label{result1}
\end{figure}

\begin{figure}[htbp]
\centering
\includegraphics[width=8.5cm]{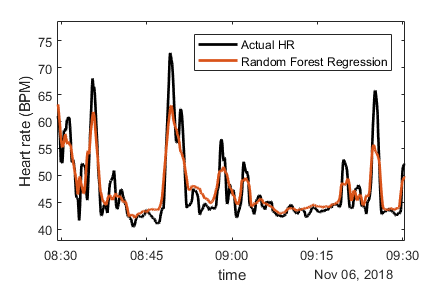}
\caption{Heart rate prediction using Random Forests as compared to ground truth.}
\label{result_RF}
\end{figure}

2)  General Models: The results of the general regression, SVM models, and random forest are shown in Figure 4. For general models, we are estimating heart rate z-score instead of the actual heart rate (beats per minute). We recorded the r-squared values of 0.18, 0.26, 0.30, and 0.66 and the RMSE value of 1.37, 1.21, 0.81, and 0.37 for the baseline, Ridge regression, SVM, and random forest model. Similarly to the personalized models, the three predictive models outperformed the baseline regression models with Random Forests recorded as best model. We also noticed the the r-squared values recorded in the generalised models are smaller than those of the personalized models. This suggests that personalized models are better predictive of heart rate than generalized models. 

\begin{figure}[htbp]
\centering
\includegraphics[width=8.5cm]{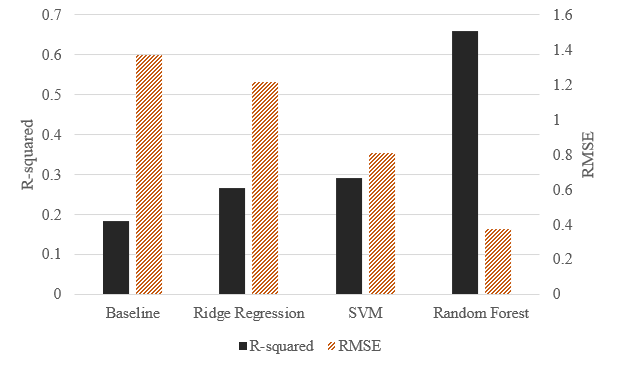}
\caption{R-squared and RMSE results of generalized ridge regression, SVM, and random forests as compared with a baseline moving-average interpolation method.}
\label{result2}
\end{figure}

\subsection{Feature Importance}
This section explores the importance of each data features used in estimating heart rate. We analyze the feature importance in the random forest model because it has the best performance as shown in Figure 2 and Figure 3. The feature importance is estimated by permuting out-of-bag observations and those estimates obtained by summing gains in the mean squared error due to splits on each feature \cite{feat1},\cite{feat2},\cite{feat3}. The random forest feature importance is shown in Figure 5. The result shows that the temporal features were the most important feature for estimating heart rate, followed by motion features, then location features. The effect of the temporal features, or the time of the day, on estimating heart rate could due to the participant's daily activity schedule such as sleeping, commuting, and work/class period, all of which can affect the person's heart rate.

\begin{figure}[htbp]
\centering
\includegraphics[width=8.5cm]{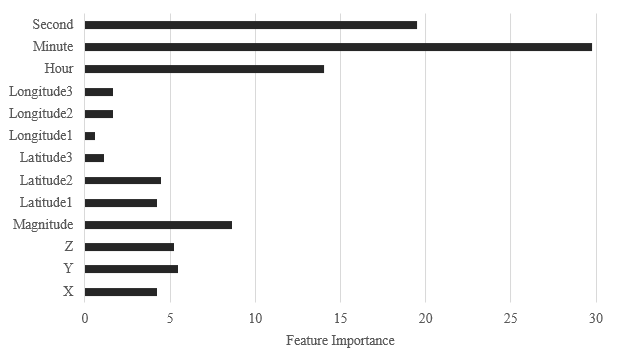}
\caption{Random forest feature importance}
\label{feature}
\end{figure}

\section{Discussion}
This work presents a framework for heart rate estimation using smartphone sensors. Although there are many limitation and confounders to the models, we believe it is a positive first step. The low R-squared values in our regression models are most likely due to two reasons; first, being the non-linear nature of our data. To combat this in the future, we would consider a Gaussian model with the heart rate being classified into deviations away from the mean heart rate. Gaussian models have been recommended for healthcare researcher \cite{b9} due to the model's ability to handle variable dependence and non-linear data while giving the researcher the ability to add knowledge about smoothness or periodicity using the covariance functions \cite{b10}. Secondly, when doing human research and real-world data collection, expected R-squared values can be lower than a controlled laboratory experiment.

The SVM and random forest regression demonstrated the best results in heart rate estimation. Even though they represent a preliminary evidence of the feasibility of predicting heart rate from smartphone sensors data, they still have some limitations. Smartphone senors might not represent participants states accurately. For instance, participants might leave the phone on table and do some physical activity, or when the phone is laying still while the participant is sleeping. Adding additional features such as light and audio to the model could help estimate heart rate when the phone is not moving. 
The random forest has the best results. However, the estimated versus actual heart rate is very close; we are concerned about overfitting (See Figure 3). Even though, random forests are utilized due to their robustness against overfitting \cite{b11}. In the future, the parameters of our random forest may need to be evaluated.

Sample size in this work is another limitation. We used a small sample of participants who all occupied the same university setting. Still, the captured data varied between participants; in the future, we consider ways, such as protocol changes, to streamline the collection and ensure the maximum amount of data is recorded. In addition, there were many collected variables we did not use for analysis, such as audio and light, that could be processed and used to improve the models. 


\section{Conclusion}
The present work investigates the feasibility of using analytical approaches to estimate heart rate data from smartphones' sensors data. Three algorithms have been implemented (ridge regression, SVM, and random forests) to estimate heart rate from multiple features extracted from time, accelerometer and GPS sensors. Both generalized and personalized models have been built and compared with baseline moving-average interpolation methods. Results demonstrate the relationship between the smartphone sensors data and heart rate and provide preliminary evidence of their predictibility of heart rate. Future work will replicate the same study in a larger population and by including other streams of data such as communication patterns (e.g. SMS and calls logs) and phone usage (e.g. Facebook usage).   








\begin{thebibliography}{00}
\bibitem{b1} W. L. Haskell, M. C. Yee, A. Evans, and P. J. Irby, “Simultaneous measurement of heart rate and body motion to quantitate physical activity.,” Med. Sci. Sports Exerc., vol. 25, no. 1, pp. 109–115, Jan. 1993.
\bibitem{b2} K. Madden and G. K. Savard, “Effects of mental state on heart rate and blood pressure variability in men and women,” Clin. Physiol., vol. 15, no. 6, pp. 557–569, 1995.
\bibitem{b3} J. M. Gorman and R. P. Sloan, “Heart rate variability in depressive and anxiety disorders,” Am. Heart J., vol. 140, no. 4, Supplement, pp. S77–S83, Oct. 2000.
\bibitem{b4} C. Schiweck, D. Piette, D. Berckmans, S. Claes, and E. Vrieze, “Heart rate and high frequency heart rate variability during stress as biomarker for clinical depression. A systematic review,” Psychol. Med., vol. 49, no. 2, pp. 200–211, Jan. 2019.
\bibitem{b5} M. T. Raggo, Mobile Data Loss: Threats and Countermeasures. Syngress, 2015.
\bibitem{b6} J. Hernandez, D. McDuff, and R. W. Picard, “Biowatch: Estimation of heart and breathing rates from wrist motions,” in 2015 9th International Conference on Pervasive Computing Technologies for Healthcare (PervasiveHealth), 2015, pp. 169–176.
\bibitem{b7} J. Hernandez, D. J. McDuff, and R. W. Picard, “Biophone: Physiology monitoring from peripheral smartphone motions,” MIT Web Domain, Aug. 2015.
\bibitem{predtHR} M. Sumida, T. Mizumoto, and K. Yasumoto, “Estimating Heart Rate Variation During Walking with Smartphone,” in Proceedings of the 2013 ACM International Joint Conference on Pervasive and Ubiquitous Computing, New York, NY, USA, 2013, pp. 245–254.
\bibitem{socialAnx} M. Boukhechba, P. Chow, K. Fua, B. A. Teachman, and L. E. Barnes, “Predicting Social Anxiety From Global Positioning System Traces of College Students: Feasibility Study,” JMIR Ment Health, vol. 5, no. 3, Jul. 2018.
\bibitem{b8} H. Xiong, Y. Huang, L. E. Barnes, and M. S. Gerber, “Sensus: A Cross-platform, General-purpose System for Mobile Crowdsensing in Human-subject Studies,” in Proceedings of the 2016 ACM International Joint Conference on Pervasive and Ubiquitous Computing, New York, NY, USA, 2016, pp. 415–426.
\bibitem{sleep} M. de Zambotti et al., “Measures of sleep and cardiac functioning during sleep using a multi-sensory commercially-available wristband in adolescents,” Physiology \& Behavior, vol. 158, pp. 143–149, May 2016.
\bibitem{feat1} L. Breiman, Classification and Regression Trees. Routledge, 2017.
\bibitem{feat2} W.-Y. Loh, “Regression Tress with Unbiased Variable Selection and Interaction Detection,” Statistica Sinica, vol. 12, no. 2, pp. 361–386, 2002.
\bibitem{feat3} W.-Y. Loh and Y.-S. Shih, “Split Selection Methods for Classification Trees,” Statistica Sinica, vol. 7, no. 4, pp. 815–840, 1997.
\bibitem{b9} J. A. Healey and R. W. Picard, “Detecting stress during real-world driving tasks using physiological sensors,” IEEE Trans. Intell. Transp. Syst., vol. 6, no. 2, pp. 156–166, Jun. 2005.
\bibitem{b10} M. K. Ameko et al., “Cluster-based Approach to Improve Affect Recognition from Passively Sensed Data,” 2018 IEEE EMBS Int. Conf. Biomed. Health Inform. BHI, pp. 434–437, Mar. 2018.
\bibitem{b11} L. Breiman, “Random Forests,” Mach. Learn., vol. 45, no. 1, pp. 5–32, Oct. 2001.
\end{thebibliography}
\end{document}